\documentclass[sigconf]{acmart}
\renewcommand\footnotetextcopyrightpermission[1]{} 

\usepackage{booktabs}
\usepackage{xr}

\copyrightyear{2019}
\acmYear{2019} 
\acmConference[WWW '19]{Proceedings of the 2019 World Wide Web Conference}{May 13--17, 2019}{San Francisco, CA, USA}
\acmDOI{10.1145/3308558.3313672}
\acmISBN{978-1-4503-6674-8/19/05}

\usepackage{balance}

\begin{document}

\title{Entity Personalized Talent Search Models \\ with Tree Interaction Features}

\author{Cagri Ozcaglar, Sahin Geyik, Brian Schmitz, Prakhar Sharma, \\ Alex Shelkovnykov, Yiming Ma, Erik Buchanan}
\affiliation{ \institution{LinkedIn} }
\email{ {caozcagl, sgeyik, bschmitz, prsharma, ashelkovnykov, yma, ebuchanan}@linkedin.com}

\begin{abstract}
Talent Search systems aim to recommend potential candidates who are a good match to the hiring needs of a recruiter expressed in terms of the recruiter's search query or job posting. Past work in this domain has focused on linear and nonlinear models which lack preference personalization in the user-level due to being trained only with globally collected recruiter activity data. In this paper, we propose an entity-personalized Talent Search model which utilizes a combination of generalized linear mixed (GLMix) models and gradient boosted decision tree (GBDT) models, and provides personalized talent recommendations using nonlinear tree interaction features generated by the GBDT. We also present the offline and online system architecture for the productionization of this hybrid model approach in our Talent Search systems. Finally, we provide offline and online experiment results benchmarking our entity-personalized model with tree interaction features, which demonstrate significant improvements in our precision metrics compared to globally trained non-personalized models.

\end{abstract}



\maketitle

\externaldocument{Background}
\externaldocument{Methodology}
\externaldocument{SystemArchitecture}
\externaldocument{ExperimentResults}
\externaldocument{Conclusion}

\section{Introduction} \label{Sec:introduction}
\let\thefootnote\relax\footnote{\\ ~ \\ ~  \large \textbf{This paper has been accepted for publication at ACM WWW 2019.}}
The task of \emph{Talent Search} is to connect job seekers with hiring managers or recruiters who are looking for candidates to fill available positions in their companies. The main way we approach this task is from the perspective of the recruiter; that is, we would like to recommend the best candidates to a recruiter who either composes an explicit search query or posts a job description. Previous work in the Talent Search domain aimed to learn a global model for the whole candidate space, by looking at the two-way interactions between recruiters and candidates, such as sending a message for an opportunity (i.e. interest from the recruiter towards the candidate, in the context of the current query or job) and receiving a positive or negative response to such messages (i.e. interest from the candidate for the opportunity) \cite{geyik_2018}, \cite{TalentSearchDeepLearning_2018}. While there has been considerable success with such methods, they still suffer because there is no personalization of preference inference for recruiters. While context of the job or query aims to contain some degree of personalization, there is often the need to learn different recommendation models at the recruiter level; so that there is a better match with the recruiter and the type of candidates that are recommended to them.

In order to personalize recommendations for different users of the same system, generalized linear mixed (GLMix) Models \cite{GLMixKDD2016}, \cite{GLMixBook2003} are often utilized, which generate per-entity linear models. In practice, GLMix models work similar to a decision tree in the way that they retrieve a different set of linear coefficients based on the id of an entity (e.g. they could employ a recruiter id $\rightarrow$ model coefficients mapping for our application case). While this introduces nonlinearity to the learned models, the features are still combined in a linear manner, and hence does not exploit the full interaction information. On the other hand, nonlinear models such as gradient boosted decision trees (GBDT) are costly to generate per entity, due to the required processing power, the size of each of the learned models, and because they are susceptible to overfitting in the case of limited data available for personalization. In this work, we propose to have the best of both worlds, by having entity-level personalization through GLMix models, while using tree interaction features \cite{PredictingClicksFB} from a globally learned GBDT model (via XGBoost \cite{XGBoostKDD2016}), utilizing the usage data of our talent search system.

The contributions of the current work are:
\begin{itemize}
\item Utilization of entity-level (i.e. recruiters and contracts) personalized models for talent search domain,
\item Achieving non-linear feature transformations within GLMix models via the tree interaction features computed by globally trained GBDT models,
\item Extensive offline experimentation of such models on recruiter recommendation data, exploring different levels of personalization (i.e. different entity-types), and,
\item Real-world online A/B test results demonstrating significant improvements in the successful search sessions, where we improve the two-way interest (both recruiter and candidates are matched within the context of an opportunity) success rate.
\end{itemize}

The organization for the rest of the paper is as follows. First, we present a brief background of the talent search domain, as well as previous personalization efforts in Section \ref{Sec:Background}, followed by the details of our methodology for personalized models in Section \ref{Sec:Methodology}. We present the offline and online architecture for the usage of personalized models in real-world talent search applications in Section \ref{Sec:SystemArchitecture}; and our offline and online experiment results in Section \ref{Sec:ExperimentResults}. Finally, we conclude the paper with a summary and discussion of future directions in Section \ref{Sec:Conclusion}.
\section{Background} \label{Sec:Background}

\subsection{Talent Search Models}
There have been several previous work that aim to solve the talent search task in different perspectives and under different constraints. One such system which generates candidates' inferred skill expertise scores using collaborative filtering based on matrix factorization, then utilizes them as features in a supervised learning-to-rank model optimizing for normalized discounted cumulative gain (NDCG, \cite{jarvelin_2002}) is presented in \cite{PersonalizedExpertiseSearchBigData2015}. An approach which aims to provide in-session personalization due to immediate user feedback is presented in \cite{geyik_2018}. We distinguish our work from this effort mainly due to the fact that we have access to the globally collected data offline to train warm-start personalized models.

An exploratory work on utilizing representation and deep learning models for talent search is presented in \cite{TalentSearchDeepLearning_2018}. There has also been a shift from querying by keyword to querying by example in talent search models \cite{QueryByExampleCIKM2017}. Finally, \emph{search by ideal candidates} approach prompts recruiters to enter ideal candidates for their open position's requirements, and a query is generated based on these ideal candidates \cite{SearchByIdealCandidatesWWW2016}. Following query generation, the talent search system retrieves and ranks results based on the match to the query.

\subsection{Personalized models}
Personalized models can be unsupervised learning models used in recommender systems, or supervised learning models, such as extensions of generalized linear models (GLMs) with user-specific model components. Recommender systems can be grouped into three categories: content-based systems, collaborative filtering systems, and latent factor models \cite{MiningMassiveDatasetsBook}. Content-based systems use the attributes of items highly rated by users, and recommend similar items to the user \cite{ContentBasedRecSystemsLops}. Collaborative filtering based systems, on the other hand, utilize user-to-user similarity based on their item preferences, and recommend the items liked by similar users \cite{MarlinCollaborativeFiltering_NIPS2003} \cite{GLSLIMRecSys2016}. Hybrid of content-based filtering and collaborative filtering methods have also been used to improve recommendations \cite{UnifyCollabContentFilteringBasilico_ICML2004} \cite{ContentBoostedCFMelville_AAAI2002} \cite{ContentBoostedCFNewsTopicRec_AAAI2015}. A third approach is latent factor models, using methods such as principal component analysis (PCA), singular value decomposition (SVD), or more recent methods such as the BellKor recommender system, which won Netflix Challenge Prize in 2009 \cite{BellKorRec2009}.

Supervised personalized models originate from standard machine learning models, with extensions in the form of additional per-entity model components. Among various methods used for prediction of users' responses to items, a common method is to use GLMs. Examples of such models are logistic regression for the case of binary response prediction, and linear regression in the case of real-valued response prediction. In use cases where the data is large and each entity has a sufficiently large dataset to make a generalization, it is possible to build a personalized model for each entity. GLMix models are an improvement over GLMs, where in addition to a global model, entity-level models are added to introduce personalization \cite{GLMixBook2003}, \cite{GLMixKDD2016}. 

\subsection{Tree Ensemble Models}
Tree-based methods partition the feature space into a set of rectangular subspaces, and then fit a model to each region \cite{ElementsOfStatisticalLearning}. Decision trees are widely used for nonlinear prediction tasks, allowing feature interactions. However, decision trees can produce unstable models which change significantly with the training data, resulting in high variance and susceptibility to overfitting. For these reasons, ensemble averaging methods such as bagging and boosting are used to reduce the variance of tree models \cite{EnsembleMethodsInMlDietterich}. Gradient Boosted Trees are tree ensemble models, where the ensemble is generated in a step-wise manner by sequentially fitting a function to residuals at each step. Each individual decision tree model is a weak learner, built by constraining the depth of the tree. Various implementations of Gradient boosted tree models have been proposed, including PLANET~\cite{PLANET}, XGBoost~\cite{XGBoostKDD2016}, parallel GBRT for web search ranking~\cite{ParallelGbrtForWebSearchRanking}, stochastic gradient boosted distributed decision trees~\cite{StochasticGradientBoostedDistributedDecisionTrees}, and PaloBoost~\cite{PaloBoost}.
\section{Methodology} \label{Sec:Methodology} 

\subsection{Generalized Linear Mixed models (GLMix)} 
GLMix models are extensions of GLMs with additional per-entity model components, and they work in the following manner for the talent search domain. Given a candidate recommendation to a recruiter uniquely identified by (requestId, contextId, recruiterId, candidateId, contractId) tuple, represented by $(r,c,re,ca,co)$, a GLMix model with per-recruiter and per-contract personalization for the Talent Search system is formulated as follows:

\begin{equation} \label{GLMixForRecruiterSearchWithMemberFeatures}
\begin{aligned}
g(\underbrace{P(r,c,re,ca,co)}_{\text{Positive Response Prob.}}) &= 
 \underbrace{\beta_{global} \cdot f_{ltr}^{r,c,re,ca,co}}_\text{Global model}  + 
 \underbrace{\beta_{re} \cdot f_{mem}^{ca}}_\text{Per-recruiter model}  \\
 & + \underbrace{\beta_{co} \cdot f_{mem}^{ca}}_\text{Per-contract model} \\
\end{aligned}
\end{equation}

\noindent where $P(r,c,re,ca,co)$ is the joint probability of recruiter $re$ sending a communication request on the position s/he would like to fill to candidate $ca$, and the candidate responding positively to it. Search query is represented by request id $r$, context $c$ (content of the query, properties of the recruiter, its company, the suggested position, etc.), and contract of the recruiter $co$ (an agreement between the recruiter and the company about the recruiting task). $g(P(r,c,re,ca,co)) = log \frac{P(r,c,re,ca,co)}{1-P(r,c,re,ca,co)}$ is the logit function. The first term on the right-hand side is the global model score with $\beta_{global}$ as the global model (i.e. fixed effect model) coefficients, the second term is the per-recruiter model score with $\beta_{re}$ as the per-recruiter model coefficients for recruiter $re$ which capture recruiter $re$'s preferences, and the third term is the per-contract model score with $\beta_{co}$ as the per-contract model coefficients for contract $co$, which capture contract $co$'s preferences. Here, feature set $f_{ltr}^{r,c,re,ca,co}$ corresponds to Learning-To-Rank (LTR) features for the search impression, uniquely identified by $(r,c,re,ca,co)$, and $f_{mem}^{ca}$ corresponds to standardized member feature vector for candidate $ca$.

The degree of entity-level personalization in GLMix models depends on the amount of data available per entity. In the Talent Search domain, for the GLMix global model + per-recruiter model + per-contract model represented in Equation \ref{GLMixForRecruiterSearchWithMemberFeatures}, there are new recruiters and contracts arriving to the system with almost no historical data, as well as recruiters and contracts who have large enough historical data for entity-level personalization. For recruiters and candidates with almost no data, the posterior means of corresponding entity-level coefficient vectors $\beta_{re}$ and $\beta_{co}$ will be close to zero, and the model for the corresponding impression uniquely identified by $(r,c,re,ca,co)$ tuple will fall back to the global model. On the other hand, for an impression with recruiter $re$ and contract $co$ with sufficient historical data, per-recruiter model with model coefficient vector $\beta_{re}$ and per-contract model with model coefficient vector $\beta_{co}$ will have impact on the final GLMix model score, which will be personalized to recruiter $re$ and contract $co$.
\subsection{Gradient Boosted Decision Trees} 
Gradient Boosted Decision Tree (GBDT) models are tree ensemble models which overcome decision trees' susceptibility to overfitting and high model variance problems via model averaging. GBDT models are computationally intensive to train. Multiple scalable implementations of GBDT model training algorithms exist in literature: \cite{PLANET} \cite{ParallelGbrtForWebSearchRanking} \cite{StochasticGradientBoostedDistributedDecisionTrees}. A commonly used GBDT model training algorithm implementation is XGBoost~\cite{XGBoostKDD2016}. XGBoost improves upon existing implementations with the following methods: ($i$) an approximate algorithm for enumerating all possible splits on all features using weighted quantile sketching, ($ii$) a sparsity-aware algorithm for parallel tree split finding and learning, ($iii$) a cache-aware block structure for out-of-core tree learning. With these improvements, XGBoost provides a highly scalable end-to-end tree boosting system.

\subsection{Best of Both Worlds: GLMix Models with Tree Interaction Features} 
In the Talent Search domain, where data is abundant, entity-level personalization via GLMix models has a high potential of improving response prediction accuracy when compared to GLM counterparts. However, GLMix models do not allow nonlinear feature interactions. GBDT models enable nonlinear feature interactions through individual decision trees and tree ensembles. On the other hand, it is infeasible to add entity-level personalization to GBDT models, due to the required processing power, size of resulting personalized models, and susceptibility to overfitting in cases of limited data availability. Therefore in this work, we use a hybrid approach: first, we employ a pre-trained GBDT model via XGBoost to generate tree interaction features and scores\footnote{~ Hyper-parameter optimization used for selecting regularization parameters for GLMix utilizes a grid search over a continuous large space for each entity-personalization level. In this work, we have preferred to use the GBDT score, instead of doing a more exhaustive hyper-parameter search with a model using tree interaction features + LTR features only, to regenerate GBDT score itself, especially helpful for global model. This reduces the computational complexity of training the GLMix model, particularly hyper-parameter optimization step.} for each recruiter search impression; next, we utilize raw features, tree interaction features, and GBDT model scores as features to build a personalized GLMix model.  

The GLMix global + per-recruiter + per-contract model with tree interaction features for a given (requestId, contextId, recruiterId, candidateId, contractId), represented by $(r,c,re,ca,co)$, can be formulated as follows:
\begin{equation}
\begin{aligned}
g(\underbrace{P(r,c,re,ca,co)}_{\text{Positive Response Prob.}}) &= 
 \underbrace{\beta_{global} \cdot f_{all}}_\text{Global model}  + 
 \underbrace{\beta_{re} \cdot f_{all}}_\text{Per-recruiter model}  \\
 & + \underbrace{\beta_{co} \cdot f_{all}}_\text{Per-contract model} \\
\end{aligned}
\end{equation}
\noindent where,
\begin{itemize}
\item	$g()$ is the logit function,
\item	$P(r,c,re,ca,co)$ is the probability of recruiter $re$ communicating with candidate $ca$ for the open position and receiving a positive response,
\item $\beta_{global}$ is the global model coefficient vector,
\item $\beta_{re}$ is the per-recruiter model coefficient vector for recruiter $re$,
\item $\beta_{co}$ is the per-contract model coefficient vector for contract $co$,
\item $f_{all} =  f_{ltr}  \bigcup f_{xgb}  \bigcup f_{int}$ is the feature vector, where $f_{ltr}$ is the set of Learning-To-Rank (LTR) features; $f_{xgb}$ represents the score from a pre-trained global GBDT model; and $f_{int}$ represents the tree interaction features from the same pre-trained global GBDT model. As an example, if \emph{feature}$_1$ is one of the LTR features, a GBDT model may generate a rule such as \emph{feature}$_1 > 0.7$ within one of the internal nodes of one of its decision trees. A tree interaction score represented by one of the leaf nodes in a decision tree of the GBDT model then is a combination of multiple rules encoded into a single rule set.
\end{itemize}
\noindent The two feature sets $f_{xgb}$ and $f_{int}$ are generated by a pre-trained GBDT model used to score the same training set. Let the training set be $D=\{x_i,y_i\}$, where $x_i \in \mathbb{R}^m$, $y_i \in \mathbb{R}$ with $|D|=n$ examples represented using $m$ features each. A GBDT model generated by XGBoost is an ensemble of tree models, which uses $K$ regression trees represented as $K$ additive functions to predict the final score of data point $x_i$:
\begin{equation}
g_{xgb}(x_i) = \displaystyle\sum_{k=1}^{K} t_k(f_{ltr}(x_i)) \textrm{,   } t_k \in T
\end{equation}
\noindent where $T=\{t(x)= w_{(q(x))}\}$ is the set of regression trees, $q:\mathbb{R}^m \rightarrow L$ represents the structure of tree $q$ which maps a data point $x_i \in \mathbb{R}^m$ to the corresponding leaf index in the tree, $w_q$ represents the leaf weight in the independent tree structure $q$, $L$ represents the number of leaves in the tree, and $f_{ltr}(x_i)$ represents LTR features of data point $x_i$. Given this representation, for each data point $x_i$, we score it using each tree $t_k$, where $k \in \{1,..,K\}$, and $x_i$ lands on leaf node $l_{i}^k$ in tree $t_k$. Each data point ends up in one leaf node of each tree in the ensemble, which encodes a set of rules chosen from the root to the leaf node of the tree. We represent these leaf nodes or tree interaction features for data point $x_i$ as a (name, term, value) triple as follows:
\begin{equation}
f_{int}(x_i) = \bigcup_{k=1}^{K}(k,l_{i}^{k},1)
\end{equation}
\noindent where $k$ is the tree index, $l_{i}^k$ is the index of the leaf node for the $k$-th tree which data point $x_i$ landed on, and binary value 1 represents that data point $x_i$ landed on this leaf node. We also use a third set of feature, $f_{xgb} (x_i)$, which is the GBDT model score:
\begin{equation}
f_{xgb} (x_i)= g_{xgb} (x_i)
\end{equation}

\begin{figure}[h]
	\includegraphics[width=3.4in]{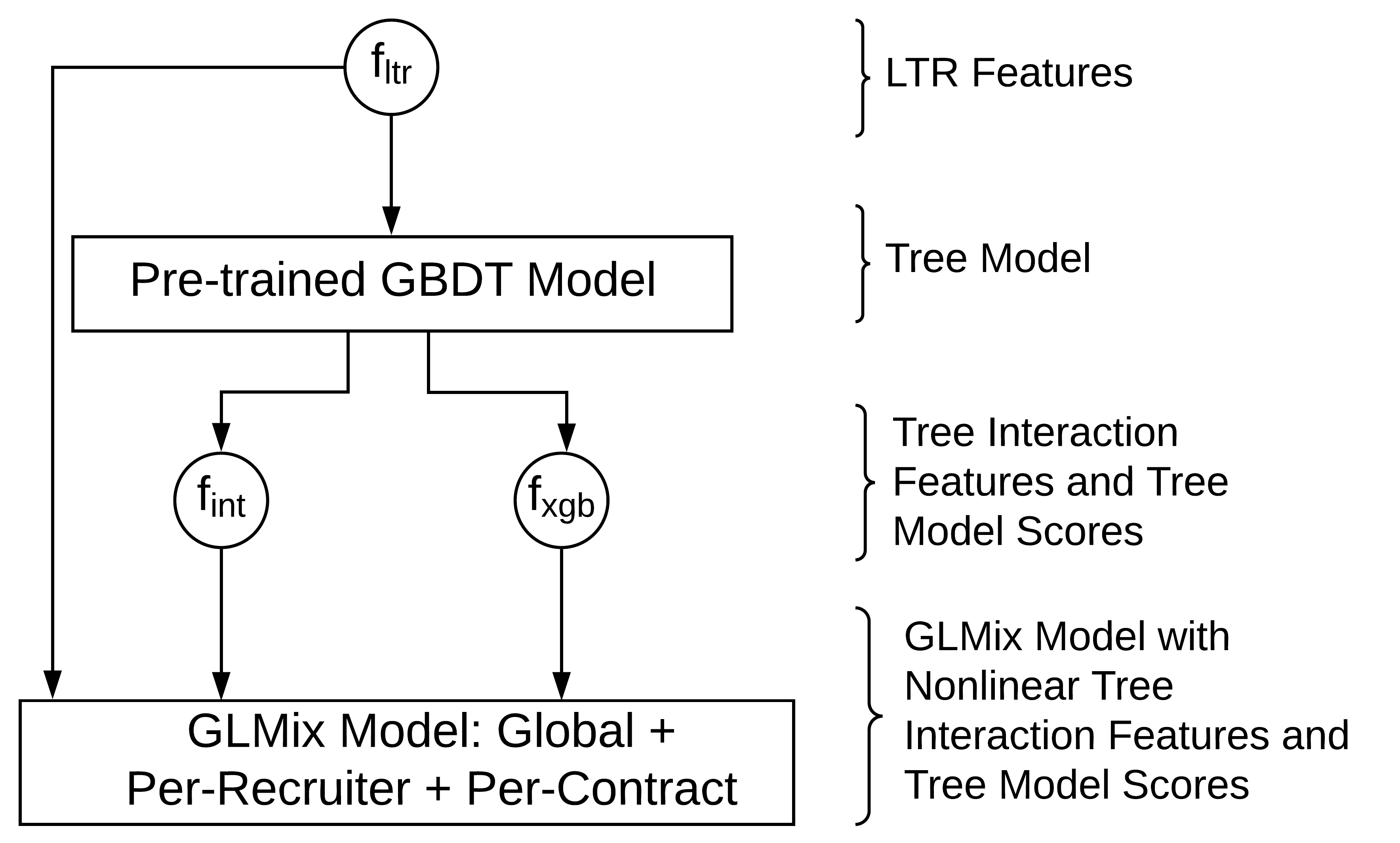}
	\vspace{-0.05in}
	\caption{Pipeline for GLMix models with nonlinear tree interaction features.}
	\label{GLMixWithTreeFeaturesFlowDiagram}
\end{figure}

\noindent For the pre-trained XGBoost model, we ran grid search on (training set, validation set) pair for (number of trees, maximum depth) hyperparameter pairs, and 100 trees with maximum depth of 2 was selected as the final hyperparameter. For the GLMix model, our hyperparemeters are the regularization weights for the global model, per-recruiter model, and per-contract model. We also ran grid search on (training set, validation set) pair for this regularization weight triple, and (100, 100, 100) was selected as the regularization weights for the global model, per-recruiter model, and per-contract model respectively.
\noindent Figure \ref{GLMixWithTreeFeaturesFlowDiagram} shows the pipeline for building GLMix models using learning-to-rank features, tree interaction features, and GBDT model scores. Learning-to-rank features are used as input to pre-trained GBDT model, which generates tree ensembles that are encoded into tree interaction features and GBDT model score for each data point. Then, using the original learning-to-rank features and their nonlinear transformations in the form of tree interaction features and GBDT model scores, we build a GLMix model with recruiter-level and contract-level personalization.

\subsection{Implementation Details}

In this section, we summarize the implementation details of the specific GLMix model and GBDT model training algorithms used, as well as the utilized system to combine those into a model training workflow. For GLMix, the open-sourced Photon-ML library \cite{PhotonMLgithubPage} was used, and for GBDTs, the open-sourced XGBoost \cite{XGBoostKDD2016} library was used. Both are built on top of the Apache Spark distributed training architecture\footnote{~ https://spark.apache.org}. Our training implementation used HDFS as the data storage platform, and YARN as the resource manager for Apache Spark.

The Photon-ML algorithm for GLMix model training works as follows. The input data is cloned to create a unique dataset for each fixed and random effect component of the GLMix model. The data in each set is identical, except for the feature vectors, which contain only the features used by that particular fixed/random effect. The data in each dataset is distributed amongst the Spark nodes used for training. The component models are trained one-by-one, starting with the fixed effect model, followed by random-effect models. After each component model is updated, residual scores are computed for all data samples using the updated model. These scores are used to offset the raw scores of successive random effect models during training. Note that the score is the value produced by a GLM prior to the application of any link function \cite{GLMixKDD2016}.

The XGBoost distributed algorithm works as follows. The input data is distributed amongst the Spark nodes in groupings known as blocks. XGBoost then runs a standard gradient tree boosting algorithm, with some modifications. Since the data is distributed, the exact greedy splitting algorithm cannot be used. Instead, XGBoost uses an approximate algorithm which proposes candidate split points according to the percentiles of feature distribution, computed using a weighted quantile sketch algorithm. The features are mapped into buckets along the proposed split points, and XGBoost chooses the best solution from amongst the proposals. XGBoost will also determine a default direction during splitting, used when feature data is missing \cite{PhotonMLgithubPage}.

In our Talent Search production system, we generate this hybrid modeling approach of entity-level personalization with tree interaction features as follows. In an offline pipeline, we generate a single day worth of training dataset using recruiter search impressions every day, hence we have a day-by-day separated dataset that accumulates over time. Then, we use 45-days of data for training, and one-day of data, immediately following the training set time period, for testing. During training / test data generation step, we also utilize a pre-trained GBDT model trained using XGBoost, in order to generate tree model score and tree interaction features for each search impression. After training and test data are generated, we build the personalized GLMix model on the 45-day training set, and test it on the one-day test set to check for and prevent extreme model performance drifts. We update the GLMix model every day, in order to personalize the model for new entities, recruiters, and contracts. Then, we upload the GLMix global model, per-recruiter model, and per-contract model coefficients into a key-value store for online utilization. More details about the online architecture of this production system are included in \textit{System Architecture} section.
\section{System Architecture} \label{Sec:SystemArchitecture}
In this section, we first present the offline pipeline we utilize to generate and store GLMix models. Then, we present the online candidate recommendation architecture, which retrieves candidates matching a user query and applies personalized ranking via a GLMix model.

\subsection{Offline Architecture}
\begin{figure}[!htb]
	\includegraphics[width=3.3in]{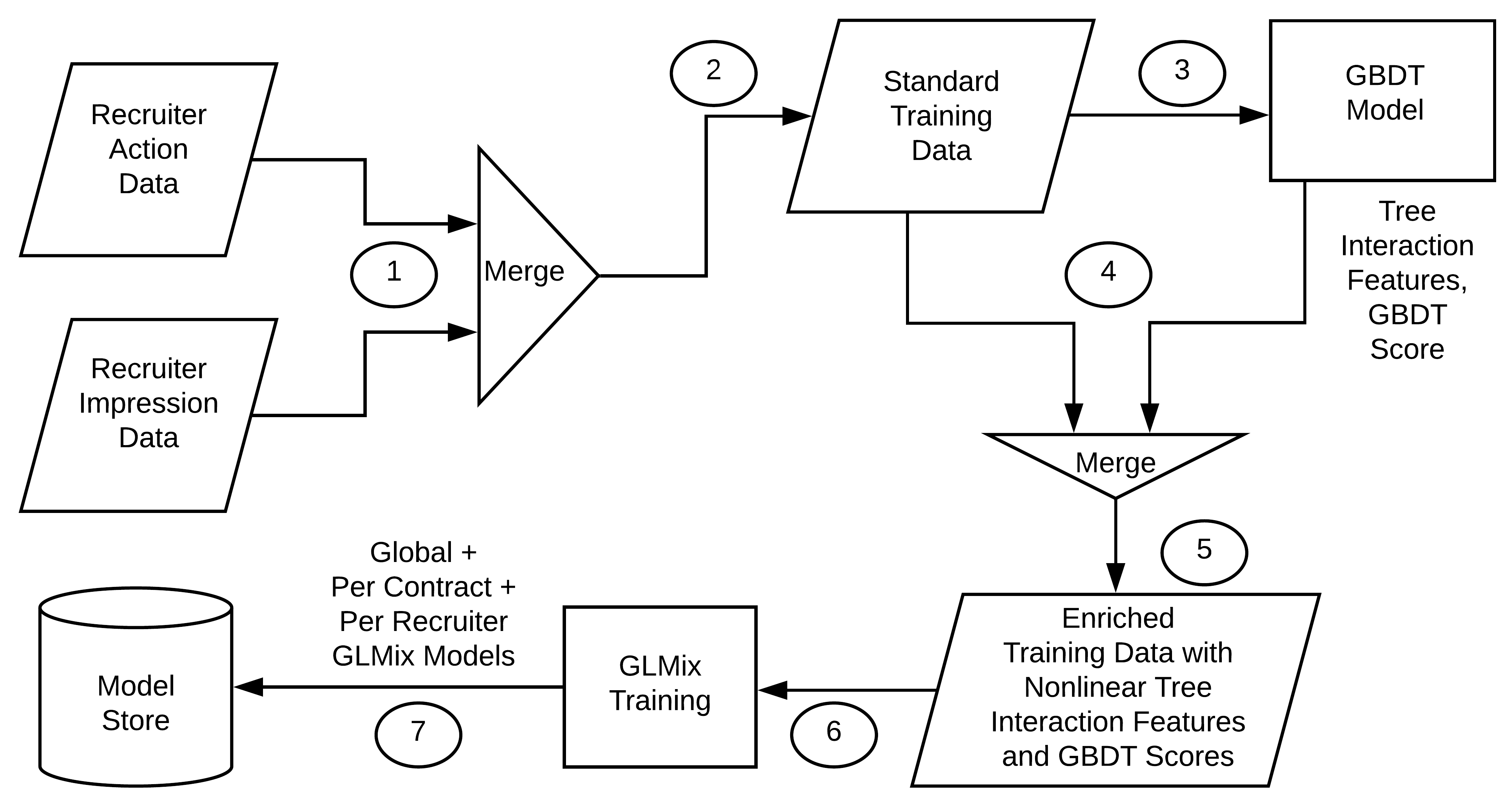}
	\vspace{-0.05in}
	\caption{Offline Architecture for generating GLMix models. The numbers denote the order in the workflow.}
	\label{OfflineArchitecture}
\end{figure}

In Figure~\ref{OfflineArchitecture}, we present the offline computation details of our layered GLMix model, previously introduced in Figure~\ref{GLMixWithTreeFeaturesFlowDiagram}. We first collect the impression data, as well as the action data from the usage of the Recruiter product. This data mainly contains the following information:
\begin{itemize}
\item Raw features collected, at the time of impression, for each candidate recommended to the users of the product\footnote{~ We are unable to reveal our raw feature list due to company policy.}.
\item Recruiter's feedback to the recommended candidates, and the candidate responses to them, as labels. This label is positive/1 if a message was sent to the candidate, and the candidate responded to the message positively. This way, we are able to model two-way interest between recruiters and candidates.
\end{itemize}
To deal with presentation bias \cite{joachims_2005}, impression and action data is collected from a subset of search query instances where we randomize the order of candidates that are recommended. Later, we feed our dataset into our pre-trained GBDT models\footnote{~ While we utilize XGBoost \cite{XGBoostKDD2016} to train the utilized GBDT models, the details of how they are trained, features as well as hyper-parameters such as depth and number of trees, are out of the scope of this paper.} \cite{geyik_2018} to generate the tree interaction features and add them to the feature vector of each training record.

We use a two-level ranking system in online architecture, which we match in the offline architecture. In the first level (L1), top K=1000 candidates are selected by model, and in the second level, a subset of 125 candidates from this L1 ranking result is selected by another model. For our hybrid GLMix model with tree interaction features, we use XGBoost model scores generated in L1, and tree interaction features generated in L2. Note that at each ranking level, available raw LTR feature set is different, that is, L1 XGBoost model generating XGBoost model score and L2 XGBoost model generating tree interaction features are different models.

Once we have our enriched dataset with raw features, tree interaction features, and XGBoost model scores, we train a GLMix model \cite{GLMixKDD2016} with global, per-recruiter, and per-contract model components (the number of personalized models is on the order of tens of thousands). The overall training process takes between 3 and 5 hours. We currently run this pipeline daily to capture the most recent recruiter and contract data, and to incorporate the latest preferences into the personalized models. Once the GLMix model is learned, the coefficients are uploaded to an online key-value store, to be used during the online serving of candidates.

\subsection{Online Architecture} 

\begin{figure}
	\includegraphics[width=3.3in]{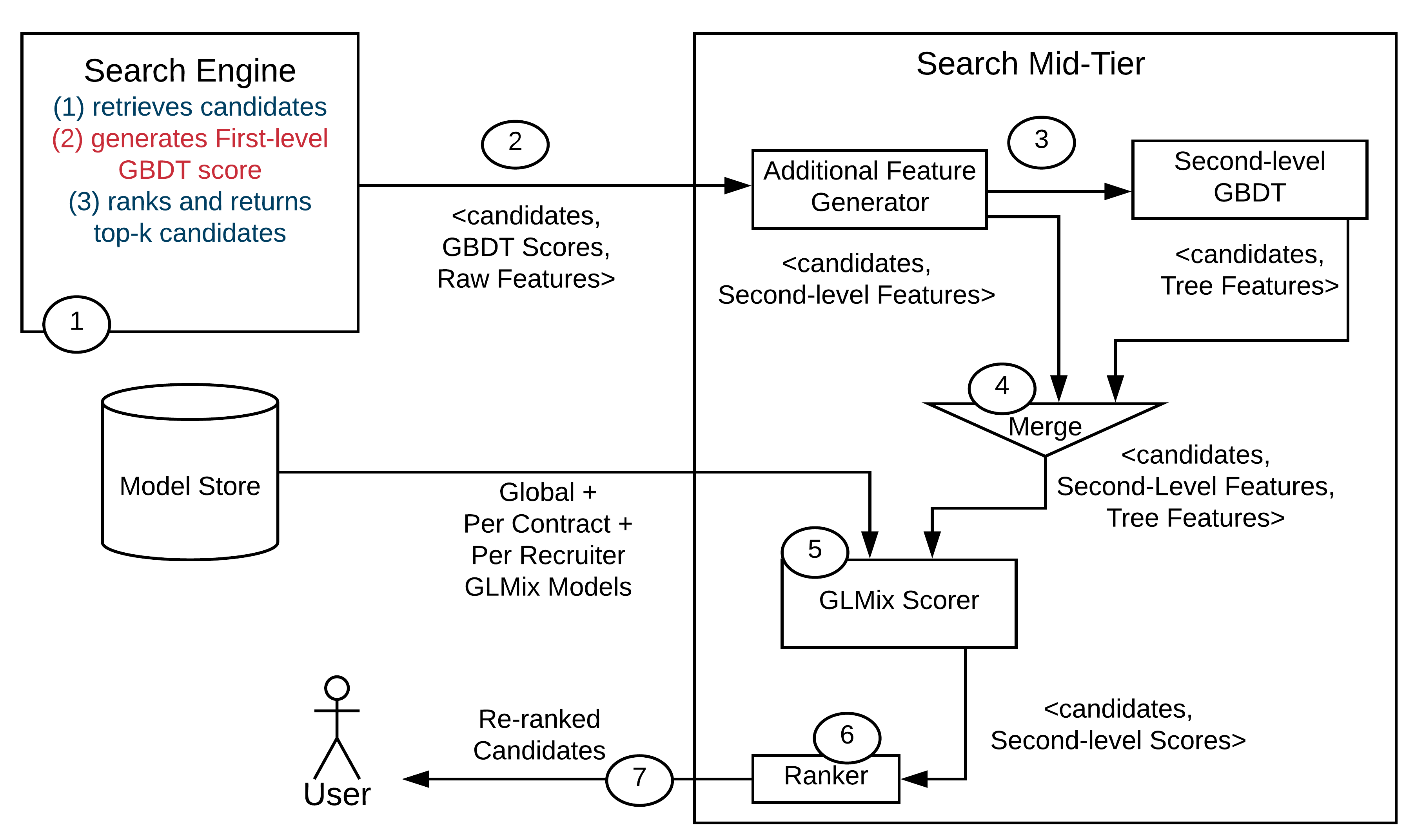}
	\vspace{-0.05in}
	\caption{Online Architecture for Ranking with GLMix models.}
	\label{OnlineArchitecture}
\end{figure}

Figure \ref{OnlineArchitecture} presents the online ranking architecture for candidates which is a two-level system. The first level retrieves and returns top-k candidates using a GBDT model. The second level utilizes our proposed GLMix model picked up from the model storage (a key-value store) to re-rank the top candidates with recruiter and contract-level personalization.

\begin{table*}[!t]
\small
\centering
\begin{tabular}{|l|l|l|l|}
\hline
\textbf{Model (Type, features)}                                  & \textbf{Precision@1} & \textbf{Precision@5} & \textbf{Precision@25} \\ \hline
Pointwise GBDT (Baseline)                                           & -       & -       & -        \\ \hline
GLMix Global, LTR + GBDT Score                                         & +7.352\%           & +3.637\%           & +0.766\%            \\ \hline
GLMix Global + Contract, LTR + GBDT Score                              & +7.688\%           & +2.609\%           & +0.683\%            \\ \hline
GLMix Global + Contract + Recruiter, LTR + GBDT Score                  & +5.429\%           & +2.267\%           & +0.683\%            \\ \hline
GLMix Global, LTR + GBDT Score \& Interaction Features                        & +4.753\%           & +3.185\%           & +1.144\%            \\ \hline
GLMix Global + Contract,  LTR + GBDT Score \& Interaction Features             & +5.909\%           & +3.941\%           & +1.594\%            \\ \hline
GLMix Global + Contract + Recruiter,  LTR + GBDT Score \& Interaction Features & +8.506\%           & +4.742\%           & +2.010\%            \\ \hline
\end{tabular}
\caption{Offline experiment results benchmarking GLMix model variants against GBDT/XGBoost model, based on \textit{Precision}@k (\textit{Positive Responses}@k) values. The values are the lifts in Precision@k values, raw Precision@k values not shown due to company policy.}
\label{OfflineExperimentResults}
\end{table*}

At query time, the search mid-tier fetches the personalized models from the key-value store and performs the following steps:
\begin{enumerate}
  \item Gets the search request from the front-end (user),
  \item Sends the query belonging to the search request into the search engine. The search engine performs the retrieval and ranking using a GBDT model (trained via XGBoost) and returns top-k candidates along with their scores and computed features. These are the first-level scores and features,
  \item Generates second-level features: There are certain features which are available only in second-level ranking due to performance constraints. Search mid-tier generates these features for the candidates returned in the first-level response,
  \item Generates tree interaction features: Tree interaction features are generated using the GBDT model in the mid-tier, and added to the second-level feature set. The model has to be the same as the one utilized in the offline pipeline to train GLMix models,
  \item Generates the second-level score using the GLMix model,
  \item Re-ranks candidates based on the second-level score, and,
  \item Sends the ranked list of candidates to the front-end (user).
\end{enumerate}

\section{Experiment Results} \label{Sec:ExperimentResults}
In order to validate the performance of GLMix models with nonlinear tree interaction features, we ran experiments in Recruiter Search domain. The key evaluation metric is precision@k, more specifically, \textit{Positive Responses}@k\footnote{~ Average number of impressions that ended in the recruiter communicating with the candidate for the opportunity, and the candidate responding positively, for the first k recommended candidates.}, for values $k \in \{1,5,25\}$, which is averaged over all search requests. In offline experiments, we benchmarked the GLMix model variants against a pointwise GBDT / XGBoost model since that was the online baseline at the time of the offline experiments. In online experiments, we benchmarked GLMix model variants against a pairwise GBDT / XGBoost model, since the baseline had moved between the original offline experimentation and online productization. Note that, other methods have been benchmarked in Talent Search domain in the past, and we use the best performing one, pointwise / pairwise XGBoost models, as baseline models in our experiments \cite{geyik_2018} \cite{TalentSearchDeepLearning_2018}.

\subsection{Offline Experiments} 
In offline experiments, we benchmarked GLMix model variants against the production model at the time, which is a pointwise GBDT/XGBoost model using learning-to-rank features. We tested GLMix model variants including:
\begin{enumerate}
\item	GLMix global model,
\item	GLMix global + per-contract model,
\item	GLMix global + per-contract + per-recruiter model.
\end{enumerate}

For GLMix model variants, we used first-level GBDT model scores and second-level GBDT model tree interaction features, optimizing for Precision@25 (the default number of results on the first page is 25 for recruiters). We employed the grid search approach for hyper-parameter selection (i.e. selection of regularization weights for global and per-entity models). Table \ref{OfflineExperimentResults} shows the comparison of Precision@k values of pointwise GBDT model and GLMix model variants. Based on these offline experiments, we observed 8.5\% / 4.7\% / 2\% lift in Precision @1 / @5 / @25 with the best-performing GLMix model variant, which is GLMix global + per-contract + per-recruiter model using LTR features, GBDT model score, and tree interaction features. Given that we optimize for Precision@25 (first page) for GLMix model when running hyper-parameter optimization, the performance of ``GLMix Global + Contract + Recruiter, LTR + GBDT Score \& Interaction Features'' model is better compared to all other GLMix variants, which is also the case for Precision@1 and Precision@5. This is also the GLMix model variant which we ran our online experiments with, and our A/B test results are presented in the next section.

\begin{table} [ht]
\small
\centering
\begin{tabular}{|c||c|c|} \hline
Metric 			& Improvement 	& p-value 	\\
 					& over Baseline	&				\\ \hline
1-day Positive	& +2.3\%			& 0.03  				\\
Response Rate	& 						&				\\ \hline
3-day Positive	&  +3\% 			& 0.01  				\\
Response Rate	& 						&				\\ \hline
7-day Positive	&  +2.7\% 			& 0.01 				\\
Response Rate	& 						&				\\ \hline
\end{tabular}
\caption{Online Results of our Three Week A/B Test}
\label{table:onlineABResults}
\end{table}

\subsection{Online Experiments}
To better understand the effect of personalization on the quality of candidates recommended to the recruiters, we have also performed real-world online tests on our LinkedIn Recruiter users, which are in the order of hundreds of thousands (and we recommend them candidates in the order of tens of millions). We are reporting the results of our A/B test which was run for around three weeks within the months of May and June 2018 in Table~\ref{table:onlineABResults}. In the presented results, 1-day positive response rate is the improvement on the percentage of communication requests sent by the recruiters that were positively responded to by the candidates (i.e. candidates were interested in the opportunity), within the first day (first three days for 3-day metric, and first seven days for 7-day metric) of sending the request. Online experiment results in this table show that GLMix model with tree interaction features results in statistically significant improvements of 2.3\%, 3\%, and 2.7\% on the 1-day, 3-day, and 7-day positive response rates, respectively, compared to the baseline pairwise GBDT/XGBoost model (as we have previously mentioned, the baseline GBDT model shifted from pointwise to pairwise during the implementation efforts for GLMix in production). Per our experience, these metrics are quite hard to move, and we can clearly see the advantages of personalization by matching the most suitable candidates based on the preferences of the recruiter and contract. The difference between the online and offline results in terms of the lifts, as presented in Tables \ref{OfflineExperimentResults} and \ref{table:onlineABResults}, is due to the change of baseline model.

\section{Conclusion and Future Work} \label{Sec:Conclusion} 

In this article, we proposed the utilization of GLMix models with tree interaction features from gradient boosted decision trees (GBDT) for Talent Search, combining the benefits of entity-level personalization and nonlinear feature interactions. We have designed an offline model training and experimentation framework, as well as an online system to employ these GLMix models in production. Our offline experiments show that the best GLMix model variant, a global + per-contract + per-recruiter model using learning-to-rank (LTR) features, GBDT model score, and GBDT model tree interaction features, outperforms the baseline pointwise GBDT model, as well as all other GLMix model variants. Our online experiments also show that the best GLMix model variant improves positive response rate with statistically significant difference over the baseline pairwise GBDT model. This method can be extended to any application in any domain where the goal is to provide personalized recommendations with nonlinear feature interactions.

For future work, one potential venue we would like to explore is to improve the hyper-parameter selection for GLMix training using Bayesian hyper-parameter optimization \cite{BayesHyperOptSnoekNIPS2012} in the daily training pipeline, compared to the currently employed grid-search approach. We also are looking into the possibility of training pairwise GLMix models instead of pointwise approach, based on our observation of high performance of pairwise GBDT models, and also due to the need to match the loss function type between our current baseline pairwise GBDT model and the GLMix model in our proposed hybrid-model solution. We also plan to replace pre-trained GBDT model with deep neural models as the nonlinear feature generator that we can feed into personalized GLMix models, which will generate wide-and-deep personalized recommendations for Talent Search.

\bibliographystyle{ACM-Reference-Format}
\balance 
\bibliography{EntityPersonalizedTalentSearchModelWithTreeInteractionFeaturesWWW}

\end{document}